%% file: manuscript_ml_uai.tex
\documentclass[accepted]{uai2023} % after acceptance, for a revised
\usepackage[british]{babel}

\usepackage{natbib} % has a nice set of citation styles and commands
\usepackage{mathtools} % amsmath with fixes and additions
\usepackage{algorithm}
\usepackage{algpseudocode}
\usepackage{booktabs} % commands to create good-looking tables
\usepackage{amsmath,amsfonts,amssymb}
\usepackage{subcaption}
\usepackage{tikz}
\usetikzlibrary{bayesnet}
\usetikzlibrary{arrows} % nice language for creating drawings and diagrams
\input{macros.tex}

\title{Massively Parallel Reweighted Wake-Sleep}

\author[1]{Thomas Heap}
\author[1]{Gavin Leech}
\author[1]{\href{mailto:<laurence.aitchison@brisol.ac.uk>?Subject=Your UAI 2023 paper}{Laurence Aitchison}{}}

\affil[1]{%
    Department of Computer Science\\
    University of Bristol\\
    Bristol
}

\begin{document}
\maketitle

\begin{abstract}
    Reweighted wake-sleep (RWS) is a machine learning method for performing Bayesian inference in a very general class of models.
    RWS draws $K$ samples from an underlying approximate posterior, then uses importance weighting to provide a better estimate of the true posterior.
    RWS then updates its approximate posterior towards the importance-weighted estimate of the true posterior.
    However, recent work \citep{chatterjee2018sample} indicates that the number of samples required for effective importance weighting is exponential in the number of latent variables.
    Attaining such a large number of importance samples is intractable in all but the smallest models.
    Here, we develop \textmp{} RWS, which circumvents this issue by drawing $K$ samples of all $n$ latent variables, and individually reasoning about all $K^n$ possible combinations of samples.
    While reasoning about $K^n$ combinations might seem intractable, the required computations can be performed in polynomial time by exploiting conditional independencies in the generative model.
    We show considerable improvements over standard ``global'' RWS, which draws $K$ samples from the full joint.
\end{abstract}

\section{Introduction}
Many machine learning tasks involve inferring the latent variables from underlying observations \citep{jaynes2003probability,mackay2003information}.
One approach to inferring these latent variables from data is to use Bayesian inference.
In Bayesian inference, we define a generative model which consists of a prior, $\P{\text{latents}}$, describing the probability of the latent variable before seeing data, and a likelihood, $\Pc{\text{data}}{\text{latents}}$, describing the probability of the data given the latents.
The goal is then to compute the posterior using Bayes theorem,
\begin{align}
  \Pc{\text{latents}}{\text{data}} \propto \Pc{\text{data}}{\text{latents}} \P{\text{latents}}.
\end{align}
However, computing this posterior is typically intractable, especially in more complex models where the likelihood or prior is parameterised by a neural network.

As an alternative, modern approaches such as  variational inference \citevi[VI; ] and reweighted wake-sleep \citerws[RWS; ] learn the parameters, $\phi$, of an approximate posterior, $\Qc[\phi]{\text{latents}}{\text{data}}$.
In VI, we learn this posterior by optimizing the evidence lower-bound objective (ELBO) using the reparameterisation trick \citep{kingma2013auto,rezende2014stochastic}.
This bound often has considerable slack, which can bias inferences.
To address this issue importance weighted auto-encoders \citeiwae[IWAEs; ] draw multiple samples from the approximate posterior and use importance weighting to provide a tighter bound on the model evidence.
In RWS, we draw multiple samples from the approximate posterior, reweight those samples to approximate the true posterior, then update the approximate posterior towards the reweighted approximation of the true posterior (specifically, this is the wake-phase Q update; see \citealp{bornschein2014reweighted}).

However, recent work \citep{chatterjee2018sample} showed that the number of samples required to get accurate importance weighted estimates is very large.
Specifically, they showed that the required number of samples scales as $e^{\Dkl{\Pc{z}{x}}{\Qc{z}{x}}}$.
This is particularly problematic because we expect the KL divergence to scale linearly in the number of latent variables, $n$.
Indeed, if $\Pc{z}{x}$ and $\Qc{z}{x}$ are IID over the $n$ latent variables, then the KL-divergence is exactly proportional to $n$.
Overall, this implies that we expect the required number of samples to be exponential in the number of latent variables, which is clearly infeasible for larger models.

This problem has been addressed in the IWAE context using TMC \citep{aitchison2019tensor}, which draws $K$ samples for each of the $n$ latent variables, and individually reasons about each of the $K^n$ combinations of samples.
Here, we develop an analogous approach for RWS, which we call \textmp{} (MP) RWS.
Critically, this is not a simple extension of the derivations in \citet{aitchison2019tensor}.
The derivations in \citet{aitchison2019tensor} are either restricted to factorised approximate posteriors, or use an augmented state-space viewpoint which cannot be applied to RWS.
We therefore give very different and considerably more general derivations in Sec.~\ref{sec:methods}.
Indeed, these more general derivations allow us to use a more general class of approximate posteriors, even in the original VI setting.

%Instead, we define \textmp{} IWAE and RWS, which draw
%It might seem that we are left again with the problem of reasoning about exponentially many ($K^n$) combinations.
%However, it turns out that we can use conditional independencies in the generative model to compute the required quantities in polynomial time.
%As such, we can implicitly obtain an exponential number of importance samples, enough to get accurate importance weighted estimates, \citep{chatterjee2018sample} in polynomial time.

\section{Related Work}
Of course, our methods are based on fundamental work on VI \citevi{}, IWAE \citeiwae{} and RWS \citerws{}.
%More specifically, there have been a few papers involving similar schemes, albeit usually in a restricted class of models.

%There is a considerable body of work on more sophisticated schemes for VI which does not apply to RWS.
Perhaps the most obvious related work is TMC \citep{aitchison2019tensor}, which also draws $K$ samples for each of the $n$ latent variables, and considers all $K^n$ combinations.
The key difference to our work is that TMC only applies to VI, while our work applies to RWS.
However, our more general derivations improve on TMC itself.
Specifically, TMC is restricted to approximate posteriors that are IID across the $K$ particles for one latent variable.
In contrast, our derivations allow us to couple the distribution over $K$ particles for a single latent variable (Appendix~\ref{app:ap}), which gives scope for e.g.\ applying variance reduction strategies.

Further, there is a body of work improving VI, but not RWS in specific restricted classes of model.

The first model class is a single-level hierarchical model, with a Bayesian parameter, $z_0$, common to all datapoints, and latent variables, $z_1\dotsc z_{n}$, each associated with a different datapoint.
\citet{geffner2022variational} propose a ``local'' importance weighting (LIW) scheme for this class of model, which contrasts with standard importance weighting schemes that they describe as ``global''.
We adopt their ``global'' terminology for standard IWAE and RWS, which draw $K$ samples from the full joint approximate posterior.
LIW in effect does IWAE separately for each datapoint: it separately draws $K$ IWAE samples for the latent variables, $z_1\dotsc z_n$, associated with each datapoint, $x_1,\dotsc,x_n$.
This looks very similar to TMC and \textmp{} RWS, which draw $K$ samples for the Bayesian parameter, $z_0$ and the latent variables, $z_1,\dotsc,z_n$, and reasons about all $K^{n+1}$ combinations of all samples on $z_0,z_1,\dotsc,z_n$.
However LIW differs from TMC and \textmp{} RWS in that LIW draws only a single sample of the Bayesian parameter, $z_0$.
Of course, there are additional differences.
In particular, LIW, like TMC, ultimately performs VI, while \textmp{} RWS applies RWS.
Further, \textmp{} RWS (like TMC) can be applied to a very broad class of models, while LIW is restricted to these single-level hierarchical models.
%As such, the differences between \textmp{} RWS and \citet{geffner2022variational} mirror those between \textmp{} and TMC.
%Specifically, both TMC and LIW are methods for VI and do not apply to RWS.
%Of course, our approach differs primarily in that it applies to RWS, whereas LIW applies to VI.
%However, there are two key additional differences between LIW and \textmp{}-like methods (including TMC).
%First, \textmp{} methods apply to a much more general class of generative model (including e.g.\ timeseries and multi-level hierarchical models amongst others).
%Second, LIW schemes draw only a single sample of $z_0$ (the latent variable common to all datapoints).
%As such, LIW in effect does IWAE separately for each datapoint; it separately draws $K$ IWAE samples for the latent variables associated with each datapoint, $z_1\dotsc z_N$.
%In contrast, \textmp{} methods also draw $K$ samples for the Bayesian parameters, $z_0$, then reason over all $K^n$ combinations.
%Differences between LIW and \textmp{} methods may be small if $z_0$ is small (e.g.\ a small dimensional vector).
%However, the differences are likely to be much larger if one tries to generalise the LIW approach to complex models with large, structured latent states.

A second class of models is timeseries models.
\textMp{} methods in timeseries may bear some resemblance to particle filtering/sequential Monte-Carlo (SMC) \citepf{}, in that SMC/particle filters also reason over multiple samples for each latent variable.
However, work which learns a proposal/approximate posterior in the particle filtering setting focuses on VI rather than RWS \citevpf{}.
Moreover, most work in SMC / particle filtering considers only a restrictive class of timeseries model, while \textmp{} methods operate in a very general class of models.
While there is some work extending SMC to more general generative models \citep[e.g.\ ][]{lindsten2017divide}, this work does not, for instance, give a mechanism to learn an approximate posterior using e.g.\ IWAE or RWS, let alone to have an approximate posterior whose structure differs from that of the underlying generative model.

\section{Background}

Here, we give background on IWAE and RWS, which are methods for performing Bayesian inference in a probabilistic generative model.
Both IWAE and RWS work with a collection of $K$ samples of the latent variables.
The full collection of $K$ samples is denoted $z$, while an individual sample (specifically the $k$th sample) is denoted $z^k$,
%\textbf{General setup.} We start with a generative model, $\P[\theta]{x, z'}$, where $x$ is our observations and $z' \in \mathcal{Z}$ is our latents.
%Our goal is to learn the parameters, $\theta$, and to approximate the posterior, $\P[\theta]{z'| x}$.
%Note that we use $z'$ rather than $z$ because we use $z$ for a collection of $K$ samples from the full latent space,
\begin{align}
  \label{eq:z}
  z &= (z^1, z^2, \dotsc, z^K) \in \mathcal{Z}^K.
\end{align}
For standard global VI and RWS, $K$ samples are drawn by sampling $K$ times from the underlying single-sample approximate posterior, $\Q[\phi]{z^k| x}$, which has parameters, $\phi$,
\begin{align}
  \label{eq:Qglobal}
  \Qglob{z| x} &= \prod_{k\in\mathcal{K}} \Q[\phi]{z^k| x},
\end{align}
where $\mathcal{K} = \{1,\dotsc,K\}$.

IWAE and RWS can be written in terms of an unbiased estimator of the marginal likelihood (Appendix~\ref{app:iwae_glob}),
%Now, we highlight and unify aspects of IWAE and RWS in a manner that will be useful for developing our approach.
%In particular, both IWAE and RWS can be written in terms of, $\mathcal{P}_\text{global}(z)$,
\begin{align}
  \label{eq:Pglobal}
  \Pglob(z) &= \frac{1}{K} \sum_{k\in\mathcal{K}} r_k(z),\\
  \label{eq:rglobal}
  %r_k(z) &= \frac{\P[\theta]{x, z^k}}{\prod_i \Q[\phi]{z_i^k| x, z_{\pa{i}}^{k}}}.
  r_k(z) &= \frac{\P[\theta]{x, z^k}}{\Q[\phi]{z^k| x}},
\end{align}
where $\P{x, z^k}$ is the generative probability, and $r_k(z)$ is the ratio of generative and approximate posterior probabilities, $r(z^k)$.

\subsection{Importance weighted autoencoder}
In IWAE \citep{burda2015importance}, we optimize $\phi$ and $\theta$ using the IWAE objective, $\Lglob$, which forms a lower-bound on the marginal likelihood, $\log \P[\theta]{x}$,
\begin{align}
  \label{eq:L:iwae}
  \log \P[\theta]{x} \geq \Lglob(\theta, \phi) &= \E[{\Qglob{z| x}}]{\log \Pglob(z)}
\end{align}
Differentiating this objective wrt the parameters of the generative model is straightforward, as $\Q[\phi]{z| x}$ does not depend on $\theta$ so we can interchange the expectation and gradient operators.
In contrast, the distribution over which the expectation is taken does depend on $\phi$, so the $\phi$ update is more difficult to implement and requires reparameterisation \citep{kingma2013auto,rezende2014stochastic}.

%this objective
%We updated $\theta$ and $\phi$ by differentiating $\mathcal{L}$,
%\begin{subequations}
%\begin{align}
%  \Delta \theta_\text{IWAE} &= \nabla_\theta \mathcal{L} = \phantom{\nabla_\phi} \E[{\Q[\phi]{z| x}}]{\nabla_\theta \log \mathcal{P}_\text{global}(z)}\\
%  \Delta \phi_\text{IWAE}   &= \nabla_\phi   \mathcal{L} = \nabla_\phi \E[{\Q[\phi]{z| x}}]{\phantom{\nabla_\theta} \log \mathcal{P}_\text{global}(z)}
%  %\E[\epsilon]{\nabla_\phi \log \mathcal{P}_\text{global}(z(\epsilon; \phi))}
%\end{align}
%\end{subequations}
%Note that the $\theta$ update is relatively easy to implement, as the distribution over which the expectation, $\Q[\phi]{z| x}$, does not depend on $\theta$, so we can put the gradient wrt $\theta$ inside the expectation.
%In contrast, the distribution over which the expectation is taken does depend on $\phi$, so the $\phi$ update is more difficult to implement.

\subsection{Reweighted wake-sleep}
In RWS \citep{bornschein2014reweighted}, we do not have a single unified objective.
%the ideal case, we would update both $\theta$ and $\phi$ using the true posterior,
%\begin{subequations}
%\begin{align}
%  \E{\Delta \theta_\text{true post}} &= \E[{\P[\theta]{z| x}}]{\nabla_\theta \log \P[\theta]{z, x}}\\
%  \E{\Delta \phi_\text{true post}} &= \E[{\P[\theta]{z| x}}]{\nabla_\phi   \log \Q[\phi]{z| x}}
%\end{align}
%\end{subequations}
%The $\theta$ update resembles the M-step in EM, while the $\phi$ update trains the approximate posterior, $\Q[\phi]{z| x}$, using maximum likelihood on samples from the true posterior, $\P[\theta]{z| x}$.
%
Instead, we update the generative model and approximate posterior by drawing $K$ samples from an approximate posterior, $\Q[\phi]{z^k| x}$.
We then use importance reweighting to bring those samples closer to the true posterior, $\P[\theta]{z| x}$, and do a maximum likelihood-like update with those reweighted samples.
In particular, the $\bareP$ update resembles the M-step in EM, and maximizes $\log \P[\theta]{z^k, x}$ for the reweighted samples.
Likewise, the (wake-phase) $\bareQ$ update maximizes $\log \Q[\phi]{z^k| x}$ for the reweighted samples mirroring the true posterior, and therefore brings $\Q[\phi]{z^k| x}$ closer to the true posterior,
\begin{subequations}
\label{eq:rws_global:iw}
\begin{align}
  \label{eq:rws_global:iw:P}
  \thetaglob &= \\
  \nonumber
  &\E[{\Qglob{z| x}}]{\frac{1}{K} \sum_{k\in\mathcal{K}} \frac{r_k(z)}{\Pglob(z)} \nabla_\theta \log \P[\theta]{z^k, x}},\\
  \label{eq:rws_global:iw:Q}
  \phiglob &= \\
  \nonumber
  &\E[{\Qglob{z| x}}]{\frac{1}{K} \sum_{k\in\mathcal{K}} \frac{r_k(z)}{\Pglob(z)} \nabla_\phi \log \Q[\phi]{z^k| x}}.
\end{align}
\end{subequations}
See Appendix~\ref{app:rws_glob} for a derivation of these updates.
However, it turns out that implementing the updates in (Eq.~\ref{eq:rws_global:iw}) directly is difficult, as it requires us to separately compute the gradients for each sample, $z^k$.
Instead, we typically use,
\begin{subequations}
\label{eq:rws_global:obj}
\begin{align}
  \label{eq:rws_global:obj:P}
  \thetaglob &= \E[{\Qglob{z| x}}]{\nabla_\theta \log \mathcal{P}_\text{global}(z)},\\
  \label{eq:rws_global:obj:Q}
  \phiglob   &= \E[{\Qglob{z| x}}]{\nabla_\phi \b{- \log \mathcal{P}_\text{global}(z)}}.
\end{align}
\end{subequations}
See Appendix~\ref{app:rws:updates} for a proof of equivalence.

\section{Methods}
\label{sec:methods}

\begin{figure*}
\begin{center}
\includegraphics[width=0.9\textwidth]{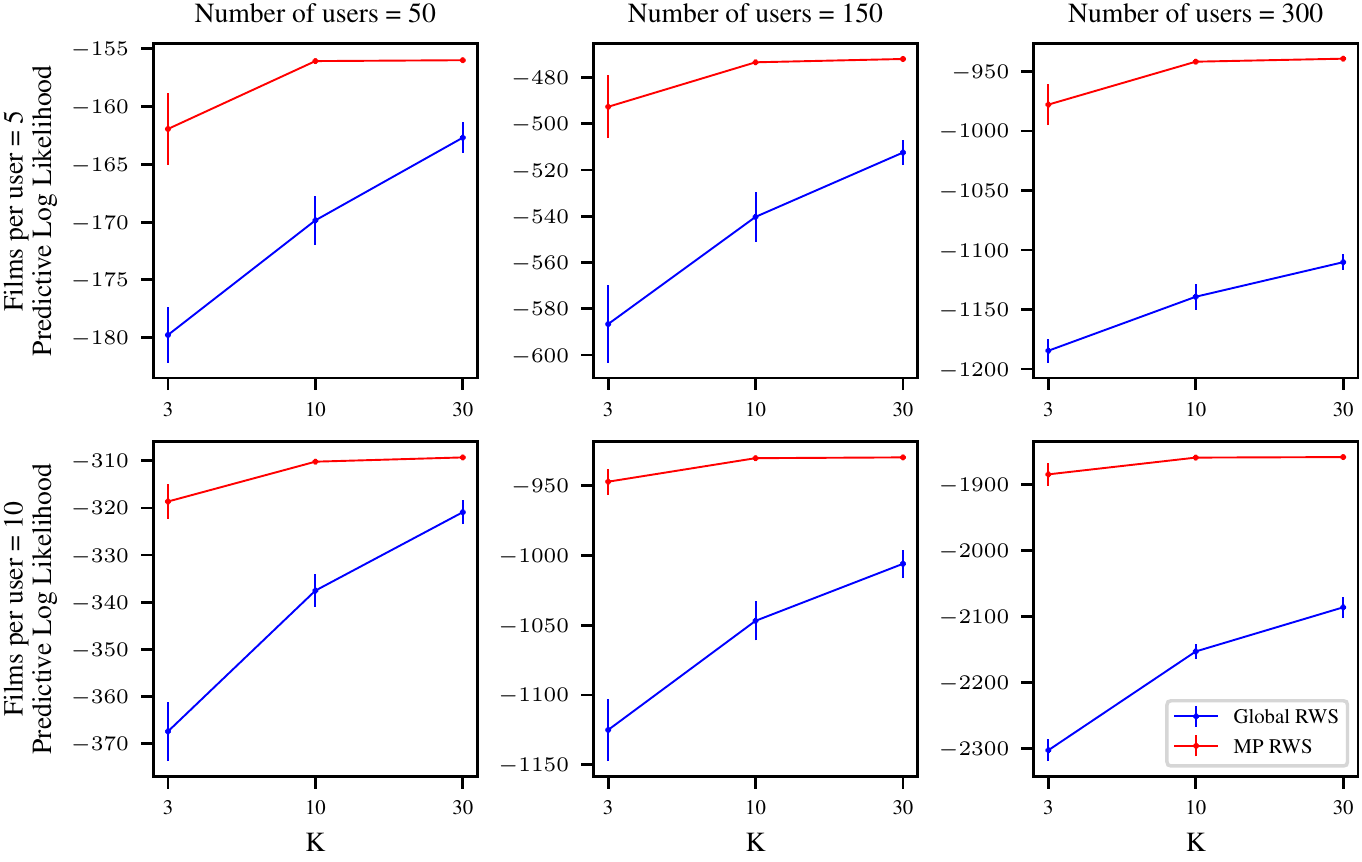}
\end{center}
\caption{Results of \textmp{} RWS and standard or ``global'' RWS for a hierarchical model on subsets of MovieLens with differing numbers of users and films per user, showing the predictive log likelihood after 25k training iterations.}
\label{fig:movielens_rws}
\end{figure*}

These previous approaches draw $K$ samples from the full joint latent space.
However, the required number of samples scales exponentially in the number of latent variables \citep{chatterjee2018sample}.
Thus, we define a \textmp{} scheme in which we draw $K$ samples for each latent variable, then effectively obtain $K^n$ samples by considering all combinations of $K$ samples for each of the $n$ latent variables.
To that end, we denote each of the separate samples for separate latent variables $z_i^k\in \mathcal{Z}_i$, where $k$ indexes the sample and $i$ indexes the latent variable.
%The $k$th sample of the full joint latent space can therefore be written,
%\begin{align}
%  z^k &= (z^k_1, z^k_2, \dotsc, z^k_n) \in \mathcal{Z}_1 \times \mathcal{Z}_2 \times \dotsm \times \mathcal{Z}_n = \mathcal{Z}.
%\end{align}
%As before (Eq.~\ref{eq:z}), we write the collection of all $K$ samples of all $n$ latent variables as $z$.
We can write the collection of $K$ samples for a single latent variable (the $i$th) as,
\begin{align}
  z_i &= (z_i^1, z_i^2, \dotsc, z_i^K) \in \mathcal{Z}_i^K.
\end{align}
To sample all $K$ copies of the full joint latent space, TMC \citep{aitchison2019tensor} uses an IID distribution over the $K$ samples, $z_i^1,\dotsc,z_i^K$,
%In contrast, TMC proposals \citep{aitchison2019tensor} were restricted to only IID distributions over the $K$ samples,
\begin{align}
  \label{eq:Qtmc}
  \Qtmcc{z}{x} &= \prod_{i=1}^n \prod_{k\in \mathcal{K}} \Qtmcc{z_i^k}{z_j \text{ for all } j \in \qa{i}}.
\end{align}
Here, $\qa{i}$ are the indices of parents of the $i$th latent variable under the approximate posterior.
In contrast, \textmp{} methods allow for dependencies between the $K$ samples for the $i$th latent variable, $z_i^1,\dotsc,z_i^K$ (Appendix~\ref{app:ap}),
\begin{align}
  %\Qc{z^k}{x} &= \prod_{i=1}^n \Qc{z_i^k}{z_{\qa{i}}^k}.\\
  %\label{eq:Qmp}
  %\Qc{z}{x} &= \prod_{i=1}^n \prod_{k\in \mathcal{K}} \Qc{z_i^k}{z_{\qa{i}}^k}
  \label{eq:Qmp}
  \Qmpc{z}{x} &= \prod_{i=1}^n \Qmpc{z_i}{z_j \text{ for all } j \in \qa{i}}.
\end{align}
There are no formal constraints on these dependencies.
However, there are practical constraints, namely that we need to be able to efficiently compute the single-particle marginals, $\Qc{z_i^k}{z_j \text{ for all } j \in \qa{i}}$.
In Appendix~\ref{app:ap}, we give more specifics about choices of $\Qmpc{z_i}{z_j \text{ for all } j \in \qa{i}}$ and $\Qtmcc{z_i^k}{z_j \text{ for all } j \in \qa{i}}$.
At a high level, these distributions are constructed by mixing the underlying single-sample approximate posterior, $\bareQ_\phi$, for different combinations of parent particles.

The generative model is more complicated, because we want to evaluate the generative probability for any of the $K^n$ possible combinations of the $K$ samples of the $n$ latent variables.
To facilitate writing down these generative probabilities, we begin by defining a vector of indices,
\begin{align}
  \k &= \b{k_1, k_2, \dotsc, k_n} \in \mathcal{K}^n,
\end{align}
which has one index, $k_i$, for each of the $n$ latent variables.
The latent variables specified by these indices is known as the ``indexed'' latent variables and can be written,
\begin{align}
  z^\k &= \b{z_1^{k_1}, z_2^{k_2}, \dotsc, z_n^{k_n}} \in \mathcal{Z}.
\end{align}
The generative probability can thus be written,
\begin{equation}
\label{eq:gen}
\begin{aligned}
  %\P[\theta]{x, z^\k} &= \P[\theta]{x| z^{\k_{\pa{x}}}_{\pa{x}}} \prod_{i=1}^n \P[\theta]{z^{k_i}_i| z^{\k_{\pa{i}}}_{\pa{i}}}.
  \P[\theta]{x, z^\k} = &\Pc[\theta]{x}{z_j^{k_j} \text{ for all } j \in \pa{x}} \\ &\prod_{i=1}^n \Pc[\theta]{z^{k_i}_i}{z_j^{k_j} \text{ for all } j \in \pa{i}}.
\end{aligned}
\end{equation}
Here, $\pa{i}$ are the indices of parents of the $i$th latent variable under the generative model, and $\pa{x}$ are the parents of the data under the generative model.
%so $z_{\pa{i}}^{\k_{\pa{i}}}$ denotes the indexed latent samples for all parents of $i$ under the generative model,
%\begin{align}
%  z_{\pa{i}}^{\k_{\pa{i}}} &= \cb{z_j^{k_j} \text{ for } j \in \pa{i}}.
%\end{align}
Our use of $\pa{i}$ mirrors our use of $\qa{i}$ to denote indices of parents of the $i$th latent variable under the approximate posterior.
Of course, the structure of the generative model and approximate posterior may differ, so $\qa{i}$ and $\pa{i}$ can also differ.

\begin{figure*}
  \centering
  \includegraphics[width=0.45\linewidth]{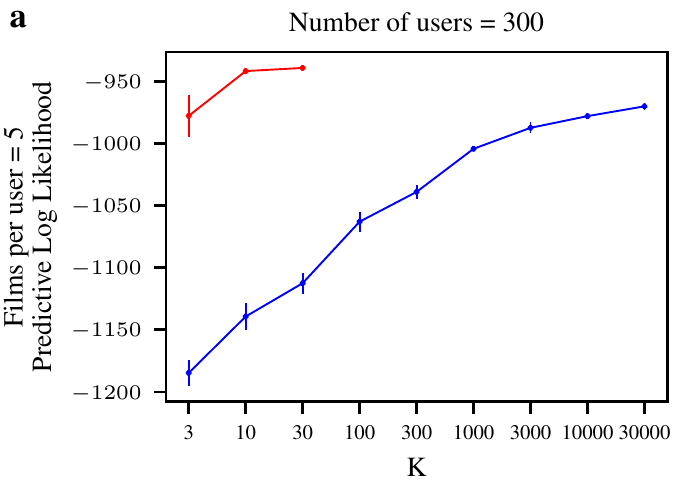}
  \includegraphics[width=0.45\linewidth]{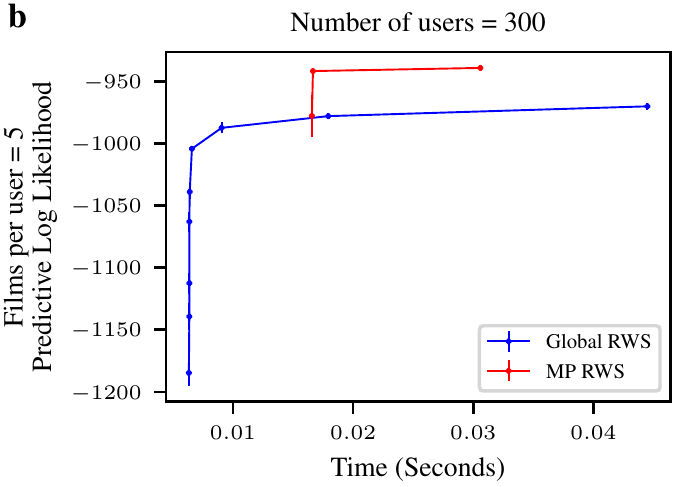}
  \caption{
    Comparison of performance between \textmp{} RWS and global RWS, for the movielens dataset. \textbf{a} Same as Fig.~\ref{fig:movielens_rws} (top right), except that we use much higher values of $K$ for global RWS.
    \textbf{b} As \textmp{} RWS may take longer than global RWS for a given $K$, we plotted time for a single training iteration against the predictive log-likelihood.
  }
 %
 % Results of inference for the movielens dataset showing predictive log likelihood after $25$k iterations. Note that Global only begins outperfoming \textmp{} RWS with $K_\mp=3$ somewhere between $K_\glob=3000$ and $K_\glob=10000$. \textbf{b} Time per single training iteration plotted against predictive log likelihood for \textmp{} RWS and globally importance sampled RWS for the Movielens dataset. Here we see that \textmp{} RWS is able to attain near maximum performance for almost no additional compute time when increasing from $K$=3 to $K$=10.
  \label{fig:movielens_comp}
\end{figure*}

%\textbf{\textmp{} IWAE and RWS.}
Looking at $\Pglob(z)$ (Eq.~\ref{eq:Pglobal}) we average only over $K$ terms, corresponding to our $K$ samples from the full joint latent space.
Our key contribution is to adapt RWS for the case where we average over all $K^n$ combinations of samples for each latent variable, indexed $\k$.

We can define an alternative unbiased marginal likelihood estimator, $\Pmp(z)$.
This estimator is obtained by averaging over all $K^n$ combinations of all samples of all latent variables,
\begin{align}
  \label{eq:Pmp}
  \Pmp(z) &= \frac{1}{K^n} \sum_{\k \in \mathcal{K}^n} r_\k(z),\\
  \label{eq:rtmc}
  r_\k(z) &= \frac{\P{x, z^\k}}{\prod_i \Qmpc{z_i^{k_i}}{z_j \text{ for all } j \in \qa{i}}}%\Q{z_i^{k_i}| x, z^{k_i}_\qa{i}}}.
\end{align}
For the proof that $\Pmp(z)$ is an unbiased marginal likelihood estimator, see Appendix~\ref{app:iwae_tmc}.
By analogy with \textglob{} IWAE, we can define an objective for \textmp{} IWAE,
\begin{align}
  \label{eq:L:tmc}
  \Lmp &= \E[{\Qmpc{z}{x}}]{\log \Pmp(z)}.
\end{align}
We prove that this quantity has the required properties (specifically, that it is a lower-bound on the log marginal likelihood) in Appendix~\ref{app:iwae_tmc}.
This quantity is very similar to that given in \citep{aitchison2019tensor}, except that it allows for a slightly more general proposal, $\bareQ_{\mp}$, which allows for dependencies between the $K$ samples for a single latent variable, $z_i^1,\dotsc,z_i^K$.
Our key contribution is to design \textmp{} updates for RWS,
\begin{subequations}
\label{eq:rws_tmc:iw}
\begin{align}
  \label{eq:rws_tmc:iw:P}
  \thetamp &{=} \E[{\Qmpc{z}{x}}]{\frac{1}{K^n}\sum_{\k\in\mathcal{K}^n} \frac{r_\k(z)}{\Pmp(z)} \nabla_\theta \log \P[\theta]{z^\k, x}}\\
  \label{eq:rws_tmc:iw:Q}
  \phimp &{=} \E[{\Qmpc{z}{x}}]{\frac{1}{K^n}\sum_{\k\in\mathcal{K}^n} \frac{r_\k(z)}{\Pmp(z)} \nabla_\phi \log \Q[\phi]{z^\k, x}}
\end{align}
\end{subequations}
These updates are derived in Appendix~\ref{app:rws_tmc}, and they can
be implemented using,
\begin{subequations}
\label{eq:rws_tmc:iw:imp}
  \begin{align}
    \label{eq:rws_tmc:iw:P:imp}
  \thetamp &= \E[{\Qmpc{z}{x}}]{\nabla_\theta \log \Pmp(z)},\\
    \label{eq:rws_tmc:iw:Q:imp}
  \phimp    &= \E[{\Qmpc{z}{x}}]{\nabla_\phi \b{- \log \Pmp(z)}},
  \end{align}
\end{subequations}
(see Appendix~\ref{app:rws_glob}).

\begin{algorithm}
\caption{Massively Parallel RWS}\label{alg:mprws}
\begin{algorithmic}
\Require{Data $x$, Prior $\mathrm{P}_\theta$, Proposal $\mathrm{Q}_\mathrm{MP}$, $K \geq 1$}

\For{$i \gets 1$ to $n$}
    \State Sample $z_i \sim \Qmpc{z_i}{z_j \text{ for all } j \in \qa{i}}$
    \State $z \gets \{z_1,...,z_{i-1}\} \cup z_i$
    \State $f^i_{k_i, \k_{\pa{i}}}(z) \gets \frac{\Pc[\theta]{z^{k_i}_i}{z_j^{k_j} \text{ for all } j \in \pa{i}}}{\Qmp{z_i^{k_i}| x, z_j \text{ for all } j \in \qa{i}}}$
\EndFor
\State $f^x_{\k_{\pa{x}}}(z) \gets \Pc[\theta]{x}{z_j^{k_j} \text{ for all } j \in \pa{x}}$
\State $\Pmp(z) \gets \tfrac{1}{K^n} \sum_{\k^n} f^x_{\k_{\pa{x}}}(z) \prod_i f^i_{k_i,\k_{\pa{i}}}(z)$
\State $\thetamp \gets {\nabla_\theta \log \Pmp(z)}$
\State $\phimp \gets {\nabla_\phi \b{- \log \Pmp(z)}}$
% \Ensure $y = x^n$
% \State $y \gets 1$
% \State $X \gets x$
% \State $N \gets n$
% \While{$N \neq 0$}
% \If{$N$ is even}
%     \State $X \gets X \times X$
%     \State $N \gets \frac{N}{2}$  \Comment{This is a comment}
% \ElsIf{$N$ is odd}
%     \State $y \gets y \times X$
%     \State $N \gets N - 1$
% \EndIf
% \EndWhile
\end{algorithmic}
\end{algorithm}

\subsection{Efficiently averaging exponentially many terms}

\begin{figure*}
  \centering
  \includegraphics[width=0.45\linewidth]{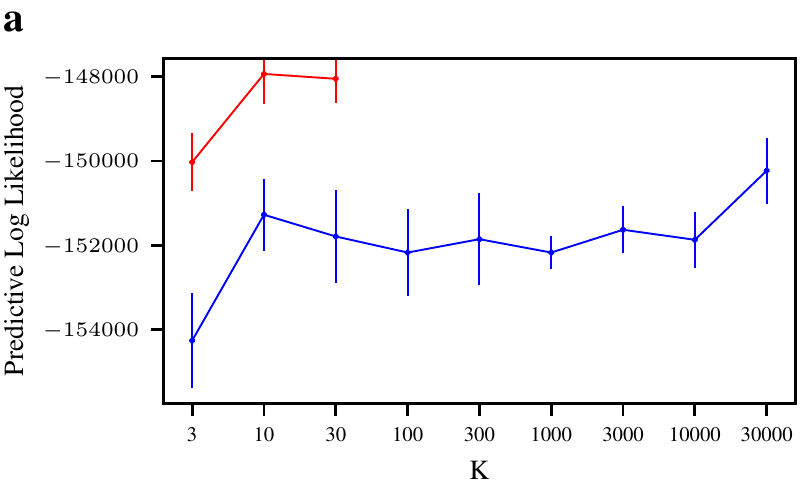}
  \includegraphics[width=0.45\linewidth]{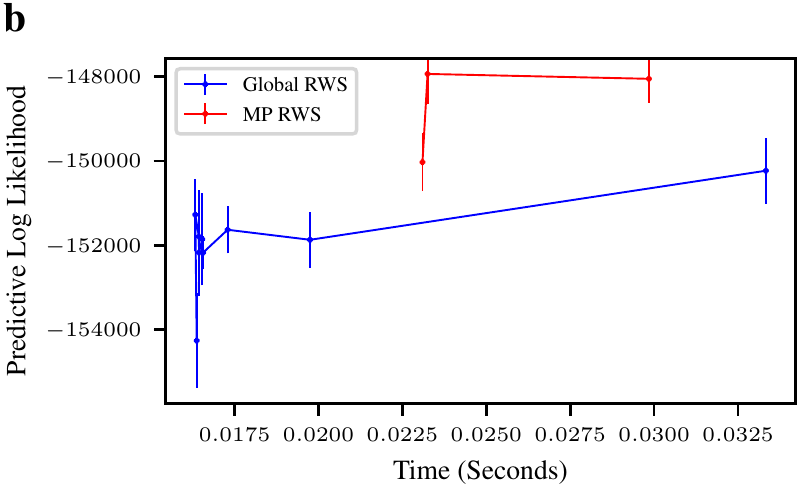}
  \caption{Comparison of performance for \textmp{} RWS and global RWS on the NYC bus breakdown dataset. \textbf{a} Predictive log-likelihood against $K$ after 75k training iterations. \textbf{b} Predictive log-likelihood against the time for a single training iteration.
  }
  \label{fig:bus_chart}
\end{figure*}

It should be surprising that we can compute $\Pmp(z)$ (Eq.~\ref{eq:Pmp}) efficiently, as it involves summing over exponentially many ($K^n$) terms.
However, it turns out that efficient computation is possible if we exploit structure in the generative model.
To exploit structure, we first need to write down the generative probability for the $\k$th sample of all latent variables, $z^\k$.
This looks alot like Eq.~\eqref{eq:gen}, as it follows the same graphical model structure, with $\pa{x}$ and $\pa{i}$ giving the indices of parents of the data and the $i$th latent variable respectively,
%; the only difference is that we need to be careful about the indices of the samples, $\k$,
%\begin{align}
  %\P[\theta]{x, z^\k} &= \P[\theta]{x| z_{\pa{x}}^{\k_\pa{x}}} \prod_{i=1}^n \P[\theta]{z^{k_i}_i| z_{\pa{i}}^{\k_\pa{i}}}.
  %\P[\theta]{x, z^\k} &= \Pc[\theta]{x}{z_j^{k_j} \text{ for all } j \in \pa{x}} \prod_{i=1}^n \Pc[\theta]{z^{k_i}_i}{z_j^{k_j} \text{ for all } j \in \pa{i}}.
%\end{align}
%However, the key difference is that we now need to be very careful about the indicies, $\k$.
%In particular, remember that the indexed latents are $z^\k = (z_1^{k_1}, z_2^{k_2}, \dotsc, z_n^{k_n})$.
%As e.g.\ the distribution over $x$ depends on only a subset of the latent variables given by $\pa{x}$, we need to be able to write the indexed latents for only the parents of $x$,
%\begin{subequations}
%\begin{align}
%  z_{\pa{x}}^{\k_{\pa{x}}} &= \b{z_j^{k_j} \text{ for all } j \in \pa{x}},\\
%  z_{\pa{i}}^{\k_{\pa{i}}} &= \b{z_j^{k_j} \text{ for all } j \in \pa{i}}.
%\end{align}
%\end{subequations}
%In particular, the indices of samples for the parent latent variables are,
%\begin{subequations}
%\begin{align}
%  \k_{\pa{x}} &= \b{k_j \text{ for all } j \in \pa{x}} \in \mathcal{K}^{\abs{\pa{x}}}, \\
%  \k_{\pa{i}} &= \b{k_j \text{ for all } j \in \pa{i}} \in \mathcal{K}^{\abs{\pa{i}}},
%\end{align}
%\end{subequations}
%where $\abs{\pa{x}}$ and $\abs{\pa{i}}$ are the number of parents for the data and the $i$th latent respectively.
%Likewise, the actual value of the $\k_{\pa{x}}$th or $\k_{\pa{i}}$th combination of the parent latent variables is,
%$\Pmp(z)$ (Eq.~\ref{eq:Pmp}),
\begin{equation}
  \label{eq:tensor_product_pq}
\begin{aligned}
  %\Pmp(z) &= \frac{1}{K^n} \sum_{\k\in\mathcal{K}^n} \P[\theta]{x| z_{\pa{x}}^{\k_\pa{x}}} \prod_{i=1}^n \frac{\P[\theta]{z^{k_i}_i| z_{\pa{i}}^{\k_\pa{i}}}}{\Q{z_i^{k_i}| x, z_\qa{i}}}
  %\Pmp(z) &= \frac{1}{K^n} \sum_{\k\in\mathcal{K}^n} \Pc[\theta]{x}{z_j^{k_j} \text{ for all } j \in \pa{x}} \prod_{i=1}^n \frac{\Pc[\theta]{z^{k_i}_i}{z_j^{k_j} \text{ for all } j \in \pa{i}}.}{\Q{z_i^{k_i}| x, z_\qa{i}}}
  r_\k(z) = &\Pc[\theta]{x}{z_j^{k_j} \text{ for all } j \in \pa{x}} \\ &\prod_{i=1}^n \frac{\Pc[\theta]{z^{k_i}_i}{z_j^{k_j} \text{ for all } j \in \pa{i}}.}{\Qmp{z_i^{k_i}| x, z_j \text{ for all } j \in \qa{i}}}
\end{aligned}
\end{equation}
If we fix $z$ (i.e. all $K^n$ samples of all $n$ latent variables), then $r_\k(z)$ can be regarded as a big tensor with $K^n$ elements, indexed by $\k$.
In that case, each term in the product defining $r_\k(z)$ (Eq.~\ref{eq:tensor_product_pq}) can also be regarded as a tensor.
The key observation is that the individual tensors in the product typically have only a few indices.
For instance, the probability of the data, $x$, depends only on the indices of samples of the parents (i.e.\ $(k_j \text{ for all } j \in \pa{x})$).
These indices of the samples of the parents can be written,
\begin{subequations}
\begin{align}
  \k_{\pa{x}} &= \b{k_j \text{ for all } j \in \pa{x}} \in \mathcal{K}^{\abs{\pa{x}}}, \\
  \k_{\pa{i}} &= \b{k_j \text{ for all } j \in \pa{i}} \in \mathcal{K}^{\abs{\pa{i}}}.
\end{align}
\end{subequations}
and where $\abs{\pa{x}}$ and $\abs{\pa{i}}$ are the number of parents latent variables.
To make explicit the idea that the individual terms in Eq.~\eqref{eq:tensor_product_pq} can be understood as tensors, we define $f^x_{\k_{\pa{x}}}(z)$ as the tensor for the data and $f^i_{k_i, \k_{\pa{i}}}(z)$ as the tensor for the $i$th latent variable,
\begin{subequations}
\begin{align}
  \label{eq:fx}
  f^x_{\k_{\pa{x}}}(z) &= \Pc[\theta]{x}{z_j^{k_j} \text{ for all } j \in \pa{x}},\\%\in \mathbb{R}^{K^{\abs{\pa{x}}}}.\\
  \label{eq:fi}
  f^i_{k_i, \k_{\pa{i}}}(z) &= \frac{\Pc[\theta]{z^{k_i}_i}{z_j^{k_j} \text{ for all } j \in \pa{i}}}{\Qmp{z_i^{k_i}| x, z_j \text{ for all } j \in \qa{i}}}.% \in \mathbb{R}^{K^{1+\abs{\pa{i}}}}.
\end{align}
\end{subequations}
Thus, we can write $r_\k(z)$ as a product of these factors,
\begin{align}
  \label{eq:tensor_product}
  r_\k(z) &= f^x_{\k_{\pa{x}}}(z) \prod_i f^i_{k_i,\k_{\pa{i}}}(z),
\end{align}
and $\Pmp(z)$ can be understood as a big tensor product,
\begin{align}
  \Pmp(z) &= \tfrac{1}{K^n} \sum_{\k^n} f^x_{\k_{\pa{x}}}(z) \prod_i f^i_{k_i,\k_{\pa{i}}}(z).
\end{align}
This tensor product can be efficiently computed in polynomial time by ordering the sums and products using Python packages such as opt-einsum \citep{daniel2018opt}.

\section{Experiments}

We present an empirical evaluation of \textmp{} RWS.
%We perform variational inference on multiple models with hierarchical latent structures.
%We compare \textmp{} RWS against TMC \cite{aitchison2019tensor} and standard importance weighted RWS.
Since the RWS wake phase $\bareQ$ requires multiple importance samples we test \textmp{} RWS (MP RWS) with $K \in \{3, 10, 30\}$ and global RWS with $K \in \{3, 10, 30,100,300,1000,3000,10000,30000\}$.
Unless otherwise stated our variational posterior is of the form $q_\phi(\mathbf{z}) = \prod_{i=1}^L q(z_{i})$, where $q(z_{i})$ is from the same family of distributions as $z_{i}$'s distribution in the generative model. We compare \textmp{} RWS (Eq.~\ref{eq:rws_tmc:iw} and Eq.~\ref{eq:rws_tmc:iw:imp}) and  against standard ``global'' RWS (Eq.~\ref{eq:rws_global:iw} and Eq.~\ref{eq:rws_global:obj}).

Optimisation is done using Adam \citep{kingma2014adam} with $\beta = (0.9, 0.999)$, no weight decay, and a learning rate of $0.001$ which is decreased by a factor of $10$ every $10$k iterations.
%The optimiser is attempting to maximise the sum of the wake phase $\phimp$ and $\thetamp$ updates.
In all cases we plot the result of 5 runs with different random seeds and plot the mean and standard error.
All times are measured on a single Nvidia A100 GPU. %except for the timeseries model where times are measured on an Core i5-7200U CPU.

\subsection{Movielens dataset}

We show results on the MovieLens100K dataset \citep{harper2015movielens}.
This dataset consists of 100K ratings from $\mathrm{M}=943$ users (indexed $m$) of $\mathrm{N}=1682$ films (indexed $j$).
Each film, indexed $j$, has as a feature vector $\mathbf{x}_j$.
We observe user ratings, and following \citep{geffner2022variational}, binarise ratings of $(0,1,2,3)$ to $0$ and ratings of $(4,5)$ to $1$.
We use the following hierarchical model:
\begin{align}
\nonumber
\m &\sim \Normal(\mathbf{0}_{18},1) \\
\nonumber
\psi &\sim \operatorname{Categorical}([0.1,0.5,0.4,0.05,0.05]) \\
\nonumber
\mathbf{z}_m &\sim \Normal(\m,\exp(\psi) \I), \ m=1,\dotsc,M \\
\label{eq:movielens_vi_rws}
\mathrm{Rating}_{mj} &\sim \operatorname{Bernoulli}(\sigma(\mathbf{z}_m^\intercal \mathbf{x}_j)), \ j=1,\dotsc,\mathrm{N}
\end{align}
This model, first samples a global mean, $\m$, and a discrete variance, $\psi$.
We then sample a vector, $\mathbf{z}_m$ for each user, which describes the types of films that they will rate highly.
The probability of a high rating is then given by taking the dot-product of the latent user-vector, $\mathbf{z}_m$ and the film's feature vector, $\mathbf{x}_j$. A corresponding graphical model can be seen in Appendix \ref{movielens}.

Note that this model has a discrete latent variable $\psi$.
As RWS does not reparameterise gradients of the ELBO, inference can proceed straightforwardly, without needing any approaches to discrete latent variables from VI, such as summing out the latent variable, applying \textsc{REINFORCE} gradient estimators or using continuous relaxations \citep{le2020revisiting}.
We compare the two methods by calculating the predictive log likelihood on a test set the same size as the training set.

To evaluate inference methods effectively, it is important to ensure that the posterior distributions are broad, and have not collapsed to very narrow point-like distributions.
As such, we evaluate on subsets of the full MovieLens dataset, composed of either 5 or 10 films per user, and 50, 150 or 300 users.

\begin{figure*}[ht]
  \centering
  \includegraphics[width=0.45\linewidth]{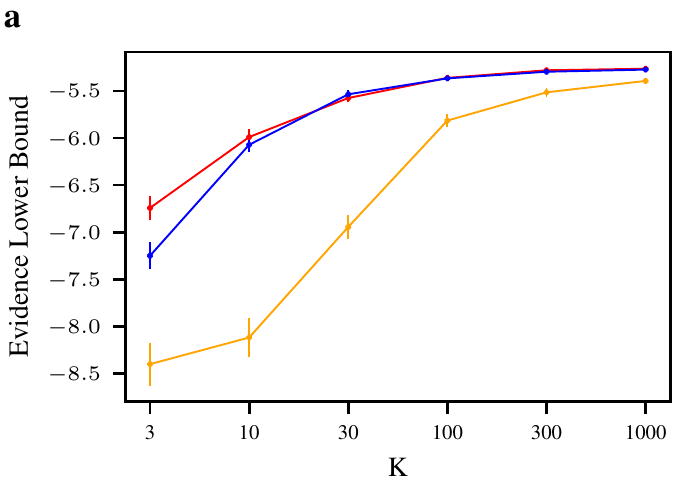}
  \includegraphics[width=0.45\linewidth]{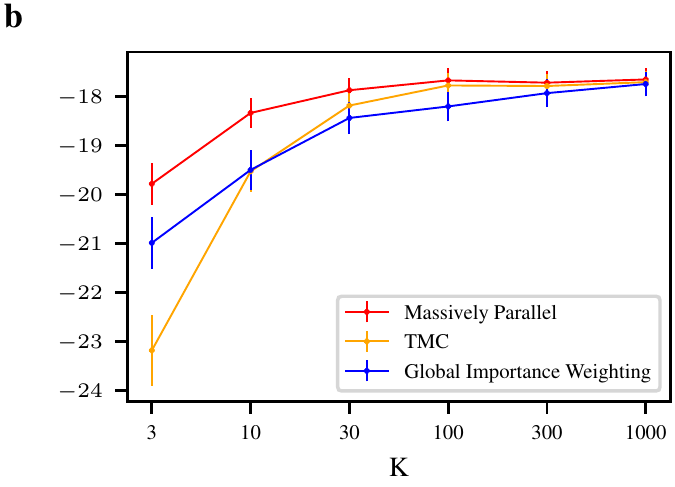}
  \caption{Comparison of performance between \textmp{} methods, TMC, particle filter and global importance weighting. \textbf{a} The timeseries model with one observation. \textbf{b} The timeseries model with multiple observations.}
  \label{fig:timeseries}
\end{figure*}
% \begin{equation}
% \begin{split}
% p(\mathbf{\mu_Z}, \mathbf{\psi_Z}, \mathbf{\psi_y}, \mathbf{y}, \mathbf{Z}) = \ & \mathcal{N}(\mathbf{\mu_Z} \vert 0, 1)
% \mathrm{Categorical}(\mathbf{\psi_Z} \vert [0.1,0.5,0.4,0.05,0.05])\\
% &\prod^M_{\mathrm{i}=1}\mathcal{N}(\mathbf{Z}_\mathrm{i} \vert \mathbf{\mu_Z}\mathbf{1}_{d_\mathbf{Z}}, \exp(\mathbf{\psi_Z})) \prod^N_{\mathrm{j}=1}\mathcal{B}(y_\mathrm{ij} \vert \mathbf{Z}^\intercal_\mathrm{i}  \mathbf{X}_\mathrm{j}))
% \end{split}
% \end{equation}

% This is the same as the model in \ref{eq:movielens_vi}, except for the discrete categorical prior over the $\mathrm{Per \ user \ mean}$ variance $\psi$.

Results are shown in Fig.~\ref{fig:movielens_rws} and Fig.~\ref{fig:movielens_comp}.
\textmp{} RWS gives considerably higher predictive log-likelihoods for all $K$ (Fig.~\ref{fig:movielens_rws},~\ref{fig:movielens_comp}a).
Importantly, the \textmp{} RWS updates are more complex than the global RWS updates, so may take longer.
We therefore also considered the performance, measured as the predictive log-likelihood, against the time for a single training iteration.
We again found considerable, albeit less dramatic, improvements (Fig.~\ref{fig:movielens_comp}b).

\subsection{NYC Bus breakdown dataset}
The city of New York releases data on the length of delays to school bus journeys \citep{nyc2023bus}. We model the length of the delay in terms of the type of journey, the school year in which the delay occurred, the borough the delay occurred in and the ID of the bus that was delayed.

To model delay time we use the model outlined in Appendix \ref{bus}. Because the dataset can be stratified into a hierarchy with three levels (Year, Borough and ID) we want our model to reflect this and, inspired by attempts to use hierarchical regression to model radon levels indoor radon levels \citep{price1996bayesian}, we use a similar multi-level regression with three levels. This model first samples a variance and mean for each year, then uses these to sample a borough mean for each year. A variance is then sampled for each borough, which together with the year level borough mean is used to sample an ID mean for each year and borough. Finally, a variance is sampled and used to sample two weight vectors, $\mathbf{C}_i$ which has length ``Number of bus companies'' and $\mathbf{J}_i$ which has length ``Number of types of journeys''. These are used to weight covariates that indicate which bus company was running a given ID's route and which type of journey was being undertaken respectively. These are then summed with the sampled ID mean for that year and borough to get the logits for a negative binomial distribution that then gives the predicted delay for the $i$-th ID in the $j$-th borough in the $m$-th year. A corresponding graphical model can be seen in Appendix \ref{bus_gm}.

We evaluate this model using a training dataset with $270$ observations: $I=30$ Ids from $J=3$ Boroughs in $M=3$ Years. We perform RWS for $75$k iterations, and evaluate the predictive log likelihood on a held out test set the same size as the training set.

Results are shown in Fig.~\ref{fig:bus_chart}. Again we see that \textmp{} RWS outperforms global RWS for all $K$.
% \begin{figure}[!h]
% \begin{center}
% \includegraphics[width=0.45\textwidth]{chart_bus.pdf}
% \end{center}
% \caption{Results of inference for the bus breakdown dataset showing predictive log likelihood after $75$k iterations.}
% \label{fig:bus_chart}
% \end{figure}

% \begin{figure}[!h]
% \begin{center}
% \includegraphics[width=0.45\textwidth]{chart_bus_time.pdf}
% \end{center}
% \caption{Time for a single training iteration plotted against predictive log likelihood for \textmp{} RWS and globally importance sampled RWS for the NYC Bus Delay dataset.}
% \label{fig:bus_chart_time}
% \end{figure}

% \begin{figure*}[ht]
%   \centering
%   \begin{subfigure}[b]{.49\textwidth}
%     \includegraphics[width=\linewidth]{chart_bus.pdf}
%     \caption{Results of inference for the bus breakdown dataset showing predictive log likelihood after $75$k iterations.}
%     \label{fig:bus_chart}
%   \end{subfigure}%
%   ~
%   \begin{subfigure}[b]{.49\textwidth}
%     \includegraphics[width=\linewidth]{chart_bus_time.pdf}
%     \caption{Time for a single training iteration plotted against predictive log likelihood for \textmp{} RWS and globally importance sampled RWS for the NYC Bus Delay dataset.}
%     \label{fig:bus_chart_time}
%   \end{subfigure}
%   \caption{Comparison of performance between \textmp{} RWS, Global RWS, for the NYC bus breakdown dataset}
%   \label{fig:bus}
% \end{figure*}

\subsection{Comparing MP VI with TMC}

Even though our main contribution is in developing \textmp{} RWS, our derivations also allow for slightly more general \textmp{} approaches to VI.
In particular, our derivations allow us to couple the proposal for the $K$ samples of the $i$th latent variable, $z_i^1,\dotsc,z_i^K$, while TMC \citep{aitchison2019tensor} forces these $K$ samples to be IID.
This coupling in \textmp{} methods allows us to introduce variance-reduction strategies inspired by methods for reducing particle degeneracy in particle filters \citedeg{} (see Appendix~\ref{app:ap} for further details).

To highlight these advantages, we considered two toy timeseries models: a single observation and a multi- observation model.
%For these timeseries models model the prior is used as a proposal. We compare both methods by computing the ELBO for $K \in \{1,3,10,30,300\}$ and plotting the average and standard error of 250 evaluations.

\subsubsection{Single Observation}
In the single observation model, there is a latent timeseries $z_1,\dotsc,z_{30}$ (we use $N=30$), and an observation, $x$, only at the last timestep,
\begin{equation}
\label{eq:timeseries}
\begin{aligned}
z_1 &= 0,\\
z_i  &\sim \Normal(z_{i-1}, 1/N),\\
x &\sim \Normal(z_N, 1)
\end{aligned}
\end{equation}
We use the prior to define the proposal (see Appendix~\ref{app:ap}).
Results can be seen in figure \ref{fig:timeseries}a.
For large $K$, all methods converge to the same value, as the ELBOs are all bounded by the true model evidence.
To compare the methods, we therefore need to consider their relative performance for smaller values of $K$.
We can see that the TMC (orange) \citep{aitchison2019tensor} performs considerably worse than \textmp{} VI (red) and IWAE (blue) \citep{burda2015importance}.
We believe that TMC is performing poorly because of particle degeneracy \citedeg{}.
In particular, the TMC proposal for $z_i$ is given by a mixture of the prior, conditioned particles from the previous timestep, $z_{i-1}$.
In sampling from this mixture, in essence, we first sample a parent particle, $z_{i-1}^{k_{i-1}}$, then we sample from the prior, conditioned on that parent sample, $\P{z_i^{k_i}| z_{i-1}^{k_{i-1}}}$.
In TMC, we choose these parent sample IID, which means that one parent particle, $z_{i-1}^{k_{i-1}}$ may have zero, one or multiple children.
This is problematic: whenever a parent sample has zero children, then this reduces diversity in the samples of $z_i$, and this issue builds up over timesteps.
\textMp{} methods circumvent this issue by ensuring that each parent sample has one and only one child sample (which requires us to couple the distribution over $z_i^1,\dotsc,z_i^K$), and IWAE avoids the issue by simply sampling $z_i^k$ conditioned on $z_{i-1}^{k}$.
\textMp{} is comparable to IWAE in this setting due to conditioning only a single scalar value at the end of the timeseries.
These methods separate when we consider multiple observations (next).

%
%We can see that \textmp{} VI (red) outperforms TMC (orange).
%on this example for all $K$, but performs comparably with global importance weighting. This pattern is expected as the benefits of \textmp{} importance weighting are not fully apparent in the case of a single observation. This is because in the case with more observations we expect the posterior to diverge further from the prior, and the larger number of ``effective'' importance weights drawn by \textmp{} methods should counterbalance this.

\subsubsection{Multiple Observations}
Next, we considered a more standard timeseries with multiple observations.
We are implementing these methods in the context of a new probabilistic programming language.
This language currently has limitations on the number of latent variables that are inherited from the opt-einsum implementation.
As such, we were not able to do the obvious thing of having one observation at every timestep.
Instead, we had an observation every third timestep.
\begin{equation}
\label{eq:timeseries more obs}
\begin{aligned}
z_1 &\sim \Normal(0, 1),\\
z_i &\sim \Normal((1-\tfrac{1}{\tau})z_{i-1}, 2/\tau), \\
x_i &\sim \Normal(z_{i}, 1) \quad\quad \text{if $i$ divisible by 3}.
\end{aligned}
\end{equation}
Again, we use $N=30$.
Results can be seen in Fig.~\ref{fig:timeseries}.
Again, the methods converge as $K$ increases, but this time, \textmp{} VI (red) gives better performance than both alternatives for lower values of $K$.
\section{Conclusion}
We introduced \textmp{} RWS, in which we draw $K$ samples for $n$ latent variables, and efficiently consider all $K^n$ combinations by exploiting conditional independencies in the generative model.
We showed that \textmp{} RWS represents a considerable improvement over previous RWS methods that draw $K$ samples from the full joint latent space.

\bibliography{refs}
% \appendix
% \onecolumn

\newpage
\onecolumn
\appendix
\title{Massively Parallel Reweighted Wake-Sleep (Supplementary Material)}
\maketitle

%% Turn this off if single column is desired for the supplement

\section{Proof of equivalence of the different forms of the global RWS updates}
\label{app:rws:updates}
We start with the RWS $\bareP$ update (Eq.~\ref{eq:rws_global:obj:P}), then use $\nabla_\theta \log \mathcal{P}_\text{global}(z) = \b{\nabla_\theta \mathcal{P}_\text{global}(z)} / \mathcal{P}_\text{global}(z)$,
\begin{align}
  % &= \E[{\Q[\phi]{z| x}}]{\nabla_\theta \log \mathcal{P}_\text{global}(z)}.\\
  %\intertext{
  \E{\Delta \theta_\text{RWS}} &= \E[{\Q[\phi]{z| x}}]{\frac{\nabla_\theta \mathcal{P}_\text{global}(z)}{\Pglob(z)}}.\\
  \intertext{Using the definition of $\mathcal{P}_\text{global}(z)$ (Eq.~\ref{eq:Pglobal}),}
  \E{\Delta \theta_\text{RWS}} &= \E[{\Q[\phi]{z| x}}]{\frac{\nabla_\theta \tfrac{1}{K} \sum_k r_k(z)}{\Pglob(z)}}\\
  \intertext{Substituting for $r_k(z)$ (Eq.~\ref{eq:rglobal}) in the numerator,}
  \E{\Delta \theta_\text{RWS}} &= \E[{\Q[\phi]{z| x}}]{\frac{\tfrac{1}{K} \sum_k \frac{\nabla_\theta \P[\theta]{z^k, x}}{\Q[\phi]{z^k| x}} }{\Pglob(z)}}\\
  \intertext{substituting $\nabla_\theta \P[\theta]{z^k, x} = \P[\theta]{z^k, x} \nabla_\theta \log \P[\theta]{z^k, x}$,}
  \E{\Delta \theta_\text{RWS}} &= \E[{\Q[\phi]{z| x}}]{\frac{1}{K} \sum_k \frac{\frac{\P[\theta]{z^k, x}}{\Q[\phi]{z^k| x}}}{\Pglob(z)}\nabla_\theta \log \P[\theta]{z^k, x}}
  %\E{\Delta \theta_\text{RWS}} &= \E[{\Q[\phi]{z| x}}]{\sum_k \frac{r(z^k)}{\sum_{k'} r(z^{k'})}\nabla_\theta \log \P[\theta]{z^k, x}},
\end{align}
Noticing that the ratio of $\P[\theta]{z^k, x}$ and $\Q[\phi]{z^k| x}$ in the numerator is equal to $r_k(z)$ (Eq.~\ref{eq:rglobal}), we get back to Eq.~\eqref{eq:rws_global:iw:P}, as required.

The RWS $\bareQ$ update is very similar.  Again, we start with Eq.~\eqref{eq:rws_global:obj:Q}, then use $\nabla_\theta \log \mathcal{P}_\text{global}(z) = \b{\nabla_\theta \mathcal{P}_\text{global}(z)} / \mathcal{P}_\text{global}(z)$,
\begin{align}
  \E{\Delta \phi_\text{RWS}}&= - \E[{\Q[\phi]{z| x}}]{\frac{\nabla_\phi \mathcal{P}_\text{global}(z)}{\mathcal{P}_\text{global}(z)}}\\
  \intertext{Using the definition of $\mathcal{P}_\text{global}(z)$ (Eq.~\ref{eq:Pglobal}),}
  \E{\Delta \phi_\text{RWS}}&= - \E[{\Q[\phi]{z| x}}]{\frac{\nabla_\phi \tfrac{1}{K} \sum_k r_k(z)}{\Pglob(z)}}\\
  \intertext{Substituting for $r_k(z)$ (Eq.~\ref{eq:rglobal}) in the numerator,}
  \E{\Delta \phi_\text{RWS}}&= - \E[{\Q[\phi]{z| x}}]{\frac{\tfrac{1}{K} \sum_k \nabla_\phi \frac{\P[\theta]{z^k, x}}{\Q[\phi]{z^k| x}} }{\Pglob(z)}}\\
  \intertext{Computing the derivative,}
  \E{\Delta \phi_\text{RWS}}&= \E[{\Q[\phi]{z| x}}]{\frac{\tfrac{1}{K} \sum_k \frac{\P[\theta]{z^k, x}}{\b{\Q[\phi]{z^k| x}}^2} \nabla_\phi \Q[\phi]{z^k| x}}{\Pglob(z)}}.\\
  \intertext{Noticing that $\b{\nabla_\phi \Q[\phi]{z^k| x}} / \Q[\phi]{z^k| x} = \nabla_\phi \log \Q[\phi]{z^k| x}$,}
  \E{\Delta \phi_\text{RWS}}&= \E[{\Q[\phi]{z| x}}]{\frac{\tfrac{1}{K} \sum_k \frac{\P[\theta]{z^k, x}}{\Q[\phi]{z^k| x}} \nabla_\phi \log \Q[\phi]{z^k| x}}{\Pglob(z)}}.
\end{align}
Finally, noticing that the ratio of $\P[\theta]{z^k, x}$ and $\Q[\phi]{z^k| x}$ in the numerator is equal to $r_k(z)$ (Eq.~\ref{eq:rglobal}), we get back to Eq.~\eqref{eq:rws_global:iw:Q}, as required.

Both of these derivations may be straightforwardly repeated for the massively parallel setting, simply by replacing $k \in \mathcal{K}$ with $\k\in\mathcal{K}^n$, and by replacing $1/K$ with $1/K^n$.

\section{TMC vs \textmp{} approximate posteriors}
\label{app:ap}
TMC approximate posteriors draw the $K$ samples of the $i$th latent variable IID,
\begin{align}
  \Qtmcc{z_i}{z_j \text{ for all } j \in \qa{i}} &=
  \prod_{k_i\in\mathcal{K}} \Qtmcc{z_i^{k_i}}{z_j \text{ for all } j \in \qa{i}}\\
  \intertext{Specifically, TMC draws each sample from an equally weighted mixture over all parent particles,}
  \Qtmcc{z_i^{k_i}}{z_j \text{ for all } j \in \qa{i}} &=
    \tfrac{1}{K^{|\qa{i}|}} \sum_{\k_{\qa{i}}} \Qglobc{z_i^{k_i}}{z_j^{k_j} \text{ for all } j \in \qa{i}}.
\end{align}
In contrast, \textmp{} methods do not force us to sample particles IID.
The key issue with IID sampling is that it introduces the risk of particle degeneracy \citedeg{}.
In particle degeneracy, some of the parent samples (e.g.\ $z_j^1$ where $j \in \qa{i}$) might have multiple children, in the sense that multiple $z_i^k$ are sampled from the mixture component arising from $z_j^1$.
At the same time, some of the parents, (e.g.\ $z_j^2$) might have no children, in the sense that no $z_i^k$ are sampled from a mixture component arising from $z_j^2$.
This is problematic because it reduces diversity in the population of samples, $z_i=(z_i^1,\dotsc,z_i^K)$, and this reduction in diversity can be especially problematic in models with long chains of latent variables, such as timeseries models.
To reduce the risk of particle degeneracy, the \textmp{} methods considered here couple the distribution over each of the $K$ particles,
\begin{align}
  \Qmpc{z_i}{z_j \text{ for all } j \in \qa{i}} &\neq \prod_{k_i\in\mathcal{K}} \Qtmcc{z_i^{k_i}}{z_j \text{ for all } j \in \qa{i}}.
\end{align}
However, we do ensure that the marginal for a single particle is the same as for TMC,
\begin{align}
  \Qmpc{z_i^{k_i}}{z_j \text{ for all } j \in \qa{i}} &=
    \tfrac{1}{K^{|\qa{i}|}} \sum_{\k_{\qa{i}}} \Qglobc{z_i^{k_i}}{z_j^{k_j} \text{ for all } j \in \qa{i}}.
\end{align}
To achieve this, we sample a permutation, $\pi$ for each latent variable, and the permutation tells us which parent particle to consider.
To give an example for one parent,
\begin{align}
  \Qmpc{z_i}{\pi, z_j} &= \prod_{k_i} \Qmpc{z_i^{k_i}}{\pi, z_j} \\
  \Qmpc{z_i^{k_i}}{\pi, z_j} &= \tfrac{1}{K} \sum_{k_i} \Qc[\phi]{z_i^{k_i}}{z_j^{\pi_{k_i}}}
\end{align}
Critically, if we marginalise over the permutation, the distribution over a single $z_i^{k_i}$ has the same density as that from a uniform mixture,
\begin{align}
  \Qmpc{z_i^{k_i}}{z_j} &= \sum_{\pi} \Qmpc{z_i^{k_i}}{\pi, z_j}\\
  \Qmpc{z_i^{k_i}}{z_j}  &= \tfrac{1}{K} \sum_{k_j} \Qc[\phi]{z_i^{k_i}}{z_j^{k_j}}.
\end{align}
Finally, if we have multiple parent latent variables, we independently sample a permutation for each latent variable.

\section{\textMp{} IWAE and RWS}
\label{app:MP}
Before getting started, it will prove useful to define some briefer notation than that used in the main text.
Specifically, we use,
\begin{align}
  z_\qa{i} &= \cb{z_j \text{ for all } j \in \qa{i}},\\
  z^{k_i}_\qa{i} &= \cb{z_j^{k_i} \text{ for all } j \in \qa{i}},\\
  z^{\k_\pa{i}}_\pa{i} &= \cb{z_j^{k_j} \text{ for all } j \in \pa{i}},
\end{align}
so,
\begin{align}
  \label{eq:Qgeneral}
  \Qc{z_i^{k_i}}{x, z_\qa{i}} &= \Qc{z_i^{k_i}}{x, z_j \text{ for all } j \in \qa{i}},\\
  % \label{eq:Qspecial}
  % \Qc{z_i^{k_i}}{x, z_\qa{i}^{k_i}} &= \Qc{z_i^{k_i}}{x, z_j^{k_i} \text{ for all } j \in \qa{i}},\\
  \Pc[\theta]{z^{k_i}_i}{z^{\k_\pa{i}}_\pa{i}} &= \Pc[\theta]{z^{k_i}_i}{z_j^{k_j} \text{ for all } j \in \pa{i}}
\end{align}
Note that in Eq.~\eqref{eq:Qgeneral}, we allow for the possibility of a slightly more general form for the approximate posterior, where the distribution over $z_i^{k_i}$ may depend on any of the parent samples.
% The usual TMC approximate posterior (Eq.~\ref{eq:Qspecial}) is a special case of this more general form.
This generalisation ensures that the subsequent derivations generalise to other possible forms for the approximate posterior, such as those for TMC (Eq.~\ref{eq:Qmp}).

In addition, it is useful to introduce notation to describe the ``non-indexed'' latent variables (i.e.\ everything in $z$ that is not $z^\k$).
The $i$th non-indexed latents are, $z_i^{/\k_i}$,
\begin{align}
  z_i^{/k_i} &= \b{z_i^{1},\dotsc,z_i^{k_i-1},z_i^{k_i+1},\dotsc,z_i^K} \in \mathcal{Z}_i^{K-1}.\\
  \intertext{and $z^{/\k}$ are all non-indexed latents,}
  z^{/\k} &= \b{z_1^{/k_1}, z_2^{/k_2}, \dotsc, z_n^{/k_n}} \in \mathcal{Z}^{K-1}.
\end{align}
%Using the notion of the non-indexed latent variables, we can write down two distributions that will become useful later.
%These distributions arise by causal interventions on the approximate posterior sampling process, \citep{pearl1995causal,pearl2000models,pearl2009causal,hernan2010causal,glymour2016causal},
%\begin{subequations}
%\label{eq:Qdo}
%\begin{align}
%  \label{eq:Qdo_z/k}
%  \Qc{z^\k}{\cdo{z^{/\k}}, x} &= \prod_i \Q{z_i^{k_i}| x, z_{\qa{i}}},\\
%  \label{eq:Qdo_zk}
%  \Qc{z^{/\k}}{\cdo{z^{\k}}, x} &= \prod_i \prod_{\kappa \in \mathcal{K}/k_i} \Q{z_i^{\kappa}| x, z_{\qa{i}}}.
%\end{align}
%\end{subequations}
%Here, $\mathcal{K}/k_i = \{1,\dotsc,k_i-1,k_i+1,\dotsc,K\}$ is the set of all indicies from $1$ to $K$, with $k_i$ left out.
%The resulting probabilities are subsets of terms in the full approximate posterior probability (Eq.~\ref{eq:Qmp}).
%Of course, distributions resulting from these causal interventions are very different from those resulting from traditional conditioning.
%For our purposes, the critical property of these two distributions resulting from different causal interventions is that the product of their probabilities is the overall joint distribution,
%\begin{align}
%  \label{eq:Qz_QkQ/k}
%  \Q{z| x} &= \Qc{z^\k}{\cdo{z^{/\k}}, x} \Qc{z^{/\k}}{\cdo{z^{\k}}, x},
%\end{align}
%which arises by inspecting the definitions of these distributions in Eq.~\eqref{eq:Qdo}.

\subsection{IWAE}
\subsubsection{Single-Sample VI}
We begin by building intuition by looking at the derivation for the ELBO in the standard single-sample VAE.
We start by writing the marginal likelihood as an integral,
\begin{align}
  \P[\theta]{x} &= \int dz' \P[\theta]{x, z'}.
  \intertext{Here, we use $z'\in \mathcal{Z}$ to denote a single sample from the full joint state space; we use $z'$ instead of $z$ because $z$ is reserved for $K$ samples (Eq.~\ref{eq:z}). Next, we divide and multiply by the approximate posterior probability, $\Q[\phi]{z'|x}$,}
  \P[\theta]{x} &= \int dz' \Q[\phi]{z'|x} \frac{\P[\theta]{x, z'}}{\Q[\phi]{z'|x}}.
  \intertext{Now, we can rewrite the integral as an expectation under the approximate posterior,}
  \label{eq:vae:E}
  \P[\theta]{x} &= \E[{\Q[\phi]{z'|x}}]{\frac{\P[\theta]{x,z'}}{\Q[\phi]{z'|x}}}.\\
  \intertext{Now we take the logarithm on both sides and apply Jensen's inequality,}
  \label{eq:vae:jensen}
  \log \P[\theta]{x} &= \log \E[{\Q[\phi]{z'|x}}]{\frac{\P[\theta]{x,z'}}{\Q[\phi]{z'|x}}} \geq \E[{\Q[\phi]{z'|x}}]{\log \frac{\P[\theta]{x,z'}}{\Q[\phi]{z'|x}}} = \L_\text{VAE}
\end{align}
Of course, this derivation is specific to the single-sample VAE.
But we can pull out an underlying strategy that generalises to the multi-sample setting.
In particular, we first come up with an unbiased estimator of the marginal likelihood.
In our VAE, this is,
\begin{align}
  \mathcal{P}_\text{VAE}(z') &= \frac{\P[\theta]{x, z'}}{\Q[\phi]{z'|x}}\\
  \intertext{Following Eq.~\eqref{eq:vae:E} we can see that this quantity is an unbiased estimator of the marginal likelihood if $z'$ is sampled from $\Q[\phi]{z'|x}$,}
  \P[\theta]{x} &= \E[{\Q[\phi]{z'|x}}]{\mathcal{P}_\text{VAE}(z')}
  \intertext{Then we apply Jensen's inequality (Eq.~\ref{eq:vae:jensen}),}
  \log \P[\theta]{x} &\geq \L_\text{VAE} = \E[{\Q[\phi]{z'|x}}]{\log \mathcal{P}_\text{VAE}(z')}.
\end{align}
However, this approach highlights key issues with the usual single-sample bound.
In particular, the single-sample estimator, $\mathcal{P}_\text{VAE}(z')$ can be very high-variance, and variance in the unbiased estimator causes the Jensen bound to become looser.

\subsubsection{Global IWAE}
\label{app:iwae_glob}
To reduce variance in the unbiased estimator, a natural approach is to average $K$ independent samples, and this is exactly what global IWAE does,
\begin{align}
  \Pglob(z) &= \frac{1}{K} \sum_{k=1}^K r_k(z) = \frac{1}{K} \sum_{k=1}^K  \mathcal{P}_\text{VAE}(z^k)
\end{align}
This is of course an unbiased estimator, as it is the average of $K$ unbiased estimators,
\begin{align}
  \P[\theta]{x} &= \E[{\Q[\phi]{z| x}}]{\Pglob(z)}.
\end{align}
Therefore, applying Jensen's inequality gives a new lower-bound on the log-marginal likelihood,
\begin{align}
  \log \P[\theta]{x} &= \log \E[{\Q[\phi]{z| x}}]{\Pmp(z)} \geq \E[{\Q[\phi]{z| x}}]{\log \Pglob(z)} = \mathcal{L}_\text{IWAE}
\end{align}
which is tighter than the usual single-sample ELBO \citep{burda2015importance}, and which matches Eq.~\eqref{eq:L:iwae} in the main text.

\subsubsection{Massively Parallel IWAE}
\label{app:iwae_tmc}
Our proposed $\Pmp(z)$ (Eq.~\ref{eq:Pmp}) is the average of $K^n$ terms, rather than $K$ terms in global IWAE.
To prove that our massively parallel strategy is valid, our strategy is to show that every term in this average is an unbiased estimator of $\log \P[\theta]{x}$, in which case the average is also an unbiased estimator, and we can again apply Jensen.

Each term in the average $\Pmp(z)$ (Eq.~\ref{eq:Pmp}) is of the form $r_\k(z)$ (Eq.~\ref{eq:rtmc}).
The expectation of each term is,
\begin{align}
  \E[{\Q[\phi]{z|x}}]{r_\k(z)} &= \E[{\Q[\phi]{z|x}}]{\frac{\P[\theta]{x, z^{\k}}}{\prod_i \Q[\phi]{z_i^{k_i}|x, z_{\qa{i}}}}}.
  \intertext{We can rewrite the expectation as an integral,}
  \E[{\Q[\phi]{z|x}}]{r_\k(z)} &= \int dz \P[\theta]{x, z^{\k}} \prod_i  \frac{\Q[\phi]{z_i| x, z_{\qa{i}}}}{\Q[\phi]{z_i^{k_i}|x, z_{\qa{i}}}}.
  \intertext{Bayes theorem tells us,}
  \frac{\Q[\phi]{z_i| x, z_{\qa{i}}}}{\Q[\phi]{z_i^{k_i}|x, z_{\qa{i}}}} &=
  \frac{\Q[\phi]{z_i^{k_i}, z_i^{/k_i}| x, z_{\qa{i}}}}{\Q[\phi]{z_i^{k_i}|x, z_{\qa{i}}}} = \Q{z_i^{/k_i}|x, z_i^{k_i}, z_{\qa{i}}},
  \intertext{Applying Bayes theorem,}
  \E[{\Q[\phi]{z|x}}]{r_\k(z)} &= \int dz \P[\theta]{x, z^{\k}} \prod_i \Q{z_i^{/k_i}|x, z_i^{k_i}, z_{\qa{i}}}.
\end{align}
Importantly, the integrand is a valid joint distribution over $x$ and $z$, or equivalently over $x$, $z^\k$ and $z^{/\k}$.
Thus, integrating over $z^{/\k}$ then $z^\k$, we find,
\begin{align}
  \E[{\Q[\phi]{z|x}}]{r_\k(z)} &= \P[\theta]{x}.
\end{align}
As such, each of the $r_\k(z)$ terms is an unbiased estimator of the marginal likelihood.
As $\Pmp(z)$ (Eq.~\ref{eq:Pmp}) is just an average of $K^n$ $r_\k(z)$ terms, it is also an unbiased estimator.
Applying Jensen's inequality to this unbiased estimator,
\begin{align}
  \log \P[\theta]{x} &\geq \E[{\Q[\phi]{z| x}}]{\log \Pmp(z)} = \Lmp,
\end{align}
which mirrors Eq.~\eqref{eq:L:tmc} in the main text.

\subsection{RWS}
\subsubsection{\textGlob{} RWS}
\label{app:rws_glob}
To build intuition, we first give a derivation of the standard RWS updates.
Ideally the updates would use samples drawn from the true posterior, $\P[\theta]{z| x}$,
\begin{subequations}
\begin{align}
  \thetapost &= \E[{\P[\theta]{z^k| x}}]{\nabla_\theta \log \P[\theta]{z, x}}\\
  \phipost &= \E[{\P[\theta]{z^k| x}}]{\nabla_\phi   \log \Q[\phi]{z| x}}
\end{align}
\end{subequations}
The $\bareP$ update is exactly the M-step in EM, and the $\bareQ$ step trains $\Q[\phi]{z|x}$ using maximum likelihood based on samples from the true posterior.
To simplify the derivations, we note that both of these updates can be understood as computing a moment under the true posterior,
\begin{align}
  \Dpost &= \E[{\P[\theta]{z^k| x}}]{\Delta(z^k)}.
\end{align}
For the $\bareP$ update, we have $\Dpost = \thetapost$ and $\Delta(z^k) = \nabla_\theta \log \P[\theta]{z, x}$.
For the $\bareQ$ update, we have $\Dpost = \phipost$ and $\Delta(z^k) = \nabla_\theta \log \Q[\phi]{z, x}$.
Of course, in practice, the true posterior is intractable, so instead we must use some form of importance weighting.
We begin by writing the generic form for the updates as an integral,
\begin{align}
  \Dpost &= \int dz^k \P{z^k| x}\Delta(z^k).
  \intertext{We then multiply and divide by an approximate posterior, $\Q{z^k| x}$,}
  \Dpost &= \int dz^k \Q{z^k| x} \frac{\P{z^k| x}}{\Q{z^k| x}} \Delta(z^k).
  \intertext{We can rewrite the integral as expectation over the approximate posterior, $\Q{z^k| x}$,}
  \Dpost &= \E[\Q{z^k| x}]{\frac{\P{z^k| x}}{\Q{z^k| x}} \Delta(z^k)}.
%  \intertext{And we can take the average over $k\in\mathcal{K}$,}
%  \Dpost &= \E[\Q{z^k| x}]{\frac{1}{K} \sum_{k\in\mathcal{K}} \frac{\P{z^k| x}}{\Q{z^k| x}} \Delta(z^k)}.
\end{align}
This quantity is difficult to use directly, because computing the posterior, $\P{z^k| x}$ involves the marginal likelihood, $\P[\theta]{x}$, which is intractable,
\begin{align}
  \P[\theta]{z^k| x} &= \tfrac{\P[\theta]{z^k, x}}{\P[\theta]{x}}  & \P[\theta]{x} = \int dz^k \P[\theta]{z^k, x}.
\end{align}
%Substituting this form for the true posterior,
%\begin{align}
%  \Dpost &= \E[\Q{z^k| x}]{\frac{\frac{\P{z^k| x}}{\Q{z^k| x}}}{\P[\theta]{x}} \Delta(z^k)}.
%\end{align}
As the true marginal likelihood is intractable, we instead use $\Pglob(z)$ (Eq.~\ref{eq:Pglobal}), which is an unbiased estimator of $\P{x}$, and is correct in the limit as $K\rightarrow \infty$ \citep{burda2015importance}.
This gives updates of the form,
\begin{align}
  \Dglob &= \E[\Q{z^k| x}]{\frac{\frac{\P{z^k, x}}{\Q{z^k| x}}}{\Pglob(z)} \Delta(z^k)}.
\end{align}
Remembering the definition of $r_k(z)$ (Eq.~\ref{eq:rglobal}), this can be written,
\begin{align}
  \Dglob &= \E[{\Q[\phi]{z|x}}]{\frac{r_k(z)}{\Pglob(z)}\Delta(z^k)}.
  %\Dglob &= \E[{\Q[\phi]{z|x}}]{\frac{1}{K} \sum_k \frac{r_k(z)}{\Pglob(z)}\Delta(z^k)}.
\end{align}
Finally as the expectation is the same for all $k$, we can average over $k$, which gives the expression in the main text (Eq.~\ref{eq:rws_global:iw})
%Finally, note that the expectation is the same for every value of $k$.
%We therefore get the same thing if we average over all $k$,
%\begin{align}
%  \Dglob &= \E[{\Q[\phi]{z|x}}]{\tfrac{1}{K} \sum_k \frac{r_k(z)}{\tfrac{1}{K} \sum_{k'} r_{k'}(z)}\Delta(z^k)}.
%\end{align}

\subsubsection{Massively Parallel RWS}
\label{app:rws_tmc}

Now, we can move on to \textmp{} RWS.
In the previous derivation for global RWS, we showed that each sample, $z^k$, individually constituted an unbiased estimator.
In the \textmp{} setting, the key difference is that instead of having $K$ samples $z^k$, we have $K^n$ samples, $z^\k$.
In particular,
\begin{align}
  \Dpost &= \E[\P{z^\k| x}]{\Delta(z^\k)} = \int dz^\k \P{z^\k| x}\Delta(z^\k).\\
  \intertext{Now, we multiply and divide by $\prod_i \Qc{z_i^{k_i}}{x, z_\qa{i}}$,}
  \label{eq:dpost_int}
  \Dpost &= \int dz^\k \b{\prod_i \Qc{z_i^{k_i}}{x, z_\qa{i}}} \frac{\P{z^\k| x}}{\prod_i \Qc{z_i^{k_i}}{x, z_\qa{i}}}  \Delta(z^\k).
\end{align}
Now, we introduce and integrate out a distribution over the non-indexed latent variables, $\prod_i \Qc{z_i^{/k_i}}{x, z_i^{k_i}, z_\qa{i}}$ \begin{align}
  \label{eq:intQz/k}
  1 &= \int dz^{/\k} \prod_i \Qc{z_i^{/k_i}}{x, z_i^{k_i}, z_\qa{i}},
\end{align}
Multiplying Eq.~\eqref{eq:dpost_int} by $1$ (Eq.~\ref{eq:intQz/k}),
\begin{align}
  \Dpost &= \int dz^\k \b{\prod_i \Qc{z_i^{k_i}}{x, z_\qa{i}}} \frac{\P{z^\k| x}}{\prod_i \Qc{z_i^{k_i}}{x, z_\qa{i}}}  \Delta(z^\k) \int dz^{/\k} \prod_i \Qc{z_i^{/k_i}}{x, z_i^{k_i}, z_\qa{i}}.
\end{align}
Combining the integrals over $z^{\k}$ and $z^{/\k}$ into a single integral over $z$, %and noting that

% \begin{align}
% \prod_i \Qc{z_i^{k_i}}{x, z_\qa{i}} \Qc{z_i^{/k_i}}{x, z_i^{k_i}, z_\qa{i}} = \Qc{z}{x},
% \end{align} %and using Eq.~\eqref{eq:Qz_QkQ/k} to combine $\Qc{z^\k}{\cdo{z^{/\k}}, x}$ and $\Qc{z^{/\k}}{\cdo{z^{\k}}, x}$ into $\Qc{z}{x}$,
\begin{align}
  \Dpost &= \int dz \Qc{z}{x} \frac{\P{z^\k| x}}{\prod_i \Qc{z_i^{k_i}}{x, z_\qa{i}}} \Delta(z^\k).
\end{align}
Writing the integral as an expectation,
\begin{align}
  \Dpost &= \E[{\Q[\phi]{z|x}}]{\frac{\P{z^\k| x}}{\prod_i \Q{z_i^{k_i}| x, z_{\pa{i}}}} \Delta(z^\k)}.
\end{align}
Again, the posterior can be written,
\begin{align}
  \P[\theta]{z^\k| x} &= \tfrac{\P[\theta]{z^\k, x}}{\P[\theta]{x}}  & \P[\theta]{x} = \int dz^\k \P[\theta]{z^\k, x}.
\end{align}
Again, the marginal likelihood, $\P[\theta]{x}$ is intractable.
Instead, we use the \textmp{} estimate of the marginal likelihood, which was shown to be unbiased in Sec.~\ref{app:iwae_tmc},
\begin{align}
  \Dmp &= \E[{\Q[\phi]{z|x}}]{\frac{\frac{\P{z^\k, x}}{\prod_i \Q{z_i^{k_i}| x, z_{\pa{i}}}}}{\Pmp(z)} \Delta(z^\k)}.
\end{align}
Remembering the definition of $r_\k(z)$ (Eq.~\ref{eq:rtmc}), this can be written,
\begin{align}
  \Dmp &= \E[{\Q[\phi]{z|x}}]{\frac{r_\k(z)}{\Pmp(z)}\Delta(z^\k)}.
\end{align}
Finally, note that the expectation is the same for every value of $\k$.
Averaging over all $K^n$ values of $\k$, we get the form in the main text (Eq.~\ref{eq:rws_tmc:iw}).

\subsection{MovieLens Graphical Model}
\label{movielens}
\begin{figure*}[!htb]
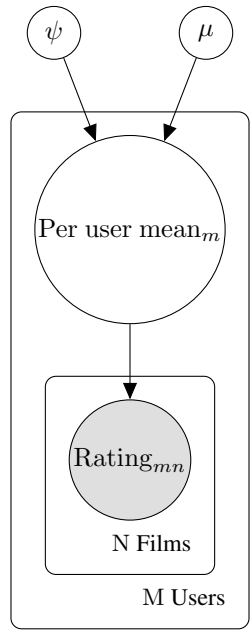

\begin{center}
  \tikz{
    % nodes
     \node[obs] (rating) {$\mathrm{Rating}_{mn}$};%
     \node[latent,above=of rating] (peruser) {$\mathrm{Per \ user \ mean}_m$}; %
     \node[latent, above=of peruser, xshift=-1cm] (psi) {$\psi$};
    \node[latent, above=of peruser, xshift=1cm] (mu) {$\mu$};
    % plate
     \plate [inner sep=.3cm] {plate2} {(rating)} {$\mathrm{N}$ Films}; %
     \plate [inner sep=.3cm] {plate1} {(peruser)(plate2)} {$\mathrm{M}$ Users}; %
    edges
     \edge {psi,mu} {peruser}
     \edge {peruser} {rating}}
\caption{Graphical model for the MovieLens dataset}
\label{fig:movielens_gm}
\end{center}
\end{figure*}

\subsection{Bus Delay Model Specification}
\label{bus}

\begin{equation}
\label{eq:bus model}
\begin{split}
\mathrm{YearVariance} &\sim \mathrm{Cat}([0.1,0.5,0.4,0.05,0.05]) \\
\mathrm{YearMean} &\sim \Normal(0,10^{-4})  \\
\mathrm{BoroughMean}_m &\sim \Normal(\mathrm{YearMean},\exp(\mathrm{YearVariance})), \ m=1,...,\mathrm{M} \\
\mathrm{BoroughVariance}_j &\sim \mathrm{Cat}([0.1,0.4,0.05,0.5,0.05]), j=1,...,\mathrm{J} \\
\mathrm{IdMean}_{mj} &\sim \Normal(\mathrm{BoroughMean}_m,\mathrm{BoroughVariance}_j), \ j=1,...,\mathrm{J}, \ m=1,...,\mathrm{M} \\
\mathrm{WeightVariance}_i &\sim \mathrm{Cat}([0.1,0.4,0.5,0.05,0.05]), \ i=1,...,\mathrm{I} \\
\mathbf{C}_i &\sim \mathcal{N}(\mathbf{0}_{\#\mathrm{BusCo.s}}, \mathrm{WeightVariance}_i), \ i=1,...,\mathrm{I} \\
\mathbf{J}_i &\sim \mathcal{N}(\mathbf{0}_{\#\mathrm{JourneyTypes}}, \mathrm{WeightVariance}_i), \ i=1,...,\mathrm{I} \\
\mathrm{logits}_{mji} &= \mathrm{IdMean}_{mj} + \mathbf{C}_i * \mathrm{Bus \ company \ name}_{mji} + \mathbf{J}_i * \mathrm{Journey \ type}_{mji} \\
\mathrm{Delay}_{mji} &\sim \mathrm{NegativeBinomial}(\mathrm{total \ count}=130, \mathrm{logits}_{mji}), \ i=1,...,\mathrm{I}, \ j=1,...,\mathrm{J}, \m=1,...,\mathrm{M}  \\
\end{split}
\end{equation}

Where $\mathrm{Bus \ company \ name}_{mji}$ is a one-hot encoded indicator variable indicating which bus company was running that route, and $\mathrm{Journey \ type}_{mji}$ similarly indicates which kind of bus journey was being undertaken. A $\mathrm{total \ county}$ of 130 is chosen as this is the largest recorded delay in the dataset.

\subsection{Bus Breakdown Graphical Model}
\label{bus_gm}
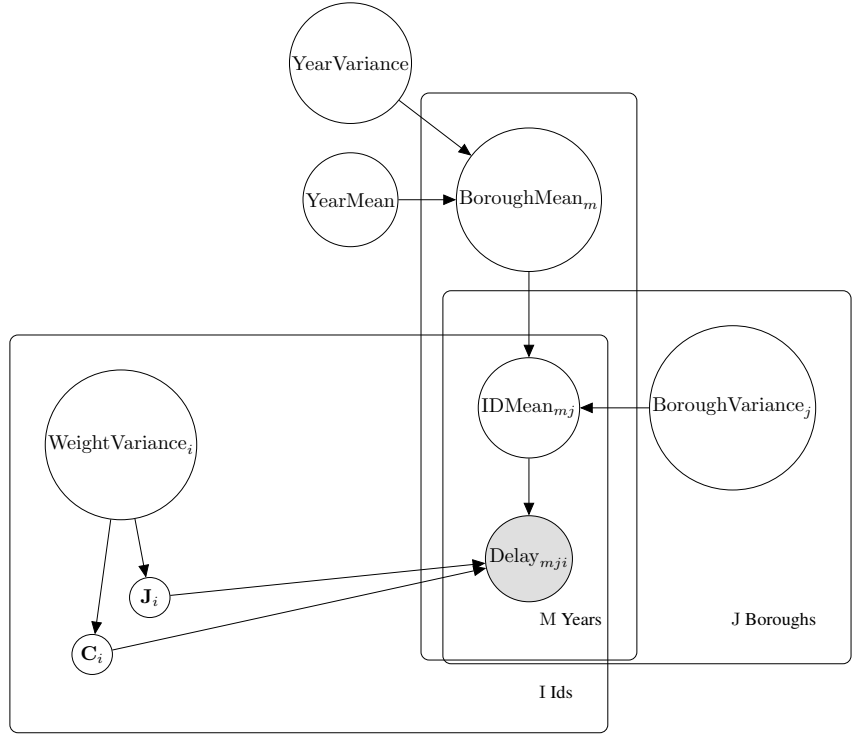
\begin{figure}[!htb]
\begin{center}
\resizebox{0.65\textwidth}{!}{%
  \begin{tikzpicture}
    % nodes

    %I
    \node[obs] (delay) {$\mathrm{Delay}_{mji}$};%

    \node[latent, left=of delay, xshift=-4cm, yshift=2cm]  (weightvariance) {$\mathrm{WeightVariance}_i$};
    \node[latent, below=of weightvariance,yshift=-1cm, xshift=-0.5cm] (companyweight) {$\mathbf{C}_i$};
    \node[latent, below=of weightvariance, xshift=0.5cm]  (journeyweight) {$\mathbf{J}_i$};

    %J
    \node[latent,above=of delay] (idmean) {$\mathrm{IDMean}_{mj}$}; %
    \node[latent, right=of idmean, xshift=0.2cm] (boroughvariance) {$\mathrm{BoroughVariance}_j$};
    %M
    \node[latent,above=of idmean, yshift=0.5cm] (boroughmean) {$\mathrm{BoroughMean}_m$}; %
    \node[latent, left=of boroughmean] (yearmean) {$\mathrm{YearMean}$};
    \node[latent, above=of yearmean, yshift=-0.5cm] (yearvariance) {$\mathrm{YearVariance}$};
    % plate
     \plate [inner sep=.6cm] {platedelay} {(delay)(companyweight)(journeyweight)(weightvariance)} {$\mathrm{I}$ Ids}; %
     \plate [inner sep=.6cm] {plateboroughs} {(idmean)(boroughvariance)(delay)} {$\mathrm{J}$ Boroughs};
     \plate [inner sep=.6cm] {plateyears} {(boroughmean)(idmean)(delay)} {$\mathrm{M}$ Years};
    %edges
     \edge {yearmean,yearvariance} {boroughmean}
     \edge {boroughmean,boroughvariance} {idmean}
     \edge {idmean} {delay}
     \edge {weightvariance} {journeyweight,companyweight}
     \edge {journeyweight,companyweight} {delay}
     \end{tikzpicture} }
\caption{Graphical model for the bus breakdown dataset}
\label{fig:bus_gm}
\end{center}
\end{figure}

\end{document}

%% file: macros.tex
\newcommand{\bracket}[3]{\left#1 #3 \right#2}
\newcommand{\mbracket}[5]{\left#1 #4 \middle#2 #5 \right#3}
\renewcommand{\b}{\bracket{(}{)}}
\newcommand{\bc}{\mbracket{(}{\vert}{)}}

\newcommand{\cb}{\bracket{\{}{\}}}
\newcommand{\abs}{\bracket{\lvert}{\rvert}}
\newcommand{\sqb}{\bracket{[}{]}}
\newcommand{\E}[1][]{\mathrm{E}_{#1}\sqb}

\newcommand{\bareP}{\operatorname{P}}
\renewcommand{\P}[1][]{\bareP_{#1}\b}
\newcommand{\Pc}[1][]{\bareP_{#1}\bc}

\newcommand{\bareQ}{{\operatorname{Q}}}

\newcommand{\Q}[1][]{\bareQ_{#1}\b}
\newcommand{\Qc}[1][]{\bareQ_{#1}\bc}
\newcommand{\Qglob}{\bareQ_{\glob}\b}
\newcommand{\Qglobc}{\bareQ_{\glob}\bc}
\newcommand{\Qmp}{\bareQ_{\mp}\b}
\newcommand{\Qmpc}{\bareQ_{\mp}\bc}

\newcommand{\Qtmcc}{\bareQ_{\tmc}\bc}

\newcommand{\I}{\mathbf{I}}

\newcommand{\m}{\boldsymbol{\mu}}

\renewcommand{\k}{\mathbf{k}}

\newcommand{\pa}[1]{{\textrm{pa}\b{#1}}}

\newcommand{\qa}[1]{{\textrm{qa}\b{#1}}}

\newcommand{\Dkl}{\operatorname{D}_\text{KL}\mbracket{(}{\Vert}{)}}

\newcommand{\tmc}{\textrm{TMC}}

\renewcommand{\mp}{\textrm{MP}}
\newcommand{\glob}{\textrm{global}}
\newcommand{\post}{\textrm{post}}

\newcommand{\Pe}{\mathcal{P}}
\renewcommand{\L}{\mathcal{L}}
\newcommand{\Pmp}{\Pe_\mp}

\newcommand{\Pglob}{\Pe_\glob}

\newcommand{\Lmp}{\L_\mp}
\newcommand{\Lglob}{\L_\glob}

\newcommand{\thetaglob}{\Delta \theta_\glob}
\newcommand{\phiglob}{\Delta \phi_\glob}
\newcommand{\thetamp}{\Delta \theta_\mp}
\newcommand{\phimp}{\Delta \phi_\mp}
\newcommand{\thetapost}{\Delta \theta_\post}
\newcommand{\phipost}{\Delta \phi_\post}

\newcommand{\Dpost}{\Delta_\post}
\newcommand{\Dglob}{\Delta_\glob}
\newcommand{\Dmp}{\Delta_\mp}

\newcommand{\textglob}{global}
\newcommand{\textGlob}{Global}

\newcommand{\textMp}{Massively parallel}
\newcommand{\textmp}{massively parallel}

\newcommand{\Normal}{\mathcal{N}}

\newcommand{\citedeg}[1][]{\citep[#1][]{carpenter1999improved,li2012deterministic,li2014fight,zhou2016new,wang2017survey}}

\newcommand{\citepf}[1][]{\citep[#1][]{gordon1993novel,doucet2009tutorial,andrieu2010particle,maddison2017filtering,le2017auto,lindsten2017divide,naesseth2018variational,lai2022variational}}

\newcommand{\citevpf}{\citep{maddison2017filtering,le2017auto,lindsten2017divide,naesseth2018variational,lai2022variational}}

\newcommand{\citevi}[1][]{\citep[#1][]{jordan1999introduction,wainwright2008graphical,kingma2013auto,rezende2014stochastic,blei2017variational,nguyen2017variational,zhang2018advances,kingma2019introduction,gayoso2021joint}}
\newcommand{\citeiwae}[1][]{\citep[#1][]{burda2015importance,cremer2017reinterpreting}}
\newcommand{\citerws}[1][]{\citep[#1][]{bornschein2014reweighted,le2020revisiting}}